%% file: main.tex
\documentclass[final]{cvpr}

\usepackage{times}
\usepackage{epsfig}
\usepackage{graphicx}
\usepackage{amsmath}
\usepackage{amssymb}
\usepackage{overpic}
\usepackage{xcolor}
\usepackage{tabu}

\usepackage{cite}

\usepackage{multicol}
\usepackage{scalerel,stackengine}
\stackMath

\input{notation}

\usepackage[pagebackref=true,breaklinks=true,colorlinks,bookmarks=false]{hyperref}

\newcommand{\PAR}[1]{\vskip4pt \noindent{\bf #1~}}
\newcommand\blfootnote[1]{%
  \begingroup
  \renewcommand\thefootnote{}\footnote{#1}%
  \addtocounter{footnote}{-1}%
  \endgroup
}

\def\cvprPaperID{1929} 
\def\confYear{CVPR 2021}

\begin{document}
\title{Human POSEitioning System (HPS): 3D Human Pose Estimation and Self-localization in Large Scenes from Body-Mounted Sensors
}
\author{Vladimir Guzov * \textsuperscript{1,2} \qquad Aymen Mir * \textsuperscript{1,2}\qquad Torsten Sattler \textsuperscript{3} \qquad Gerard Pons-Moll\textsuperscript{1,2}\\\\
{\small \textsuperscript{1}University of Tübingen,  Germany, \qquad \textsuperscript{2}Max Planck Institute for Informatics, Saarland Informatics Campus, Germany}  \\
{\small\textsuperscript{3}CIIRC, Czech Technical University in Prague, Czech Republic}\\
{\tt\scriptsize \{vguzov, amir, gpons\}@mpi-inf.mpg.de torsten.sattler@cvut.cz}}

\makeatletter
\let\@oldmaketitle\@maketitle
\renewcommand{\@maketitle}{
	\@oldmaketitle
	\begin{center}
	\vspace{-6mm}
 	\includegraphics[width=1\linewidth]{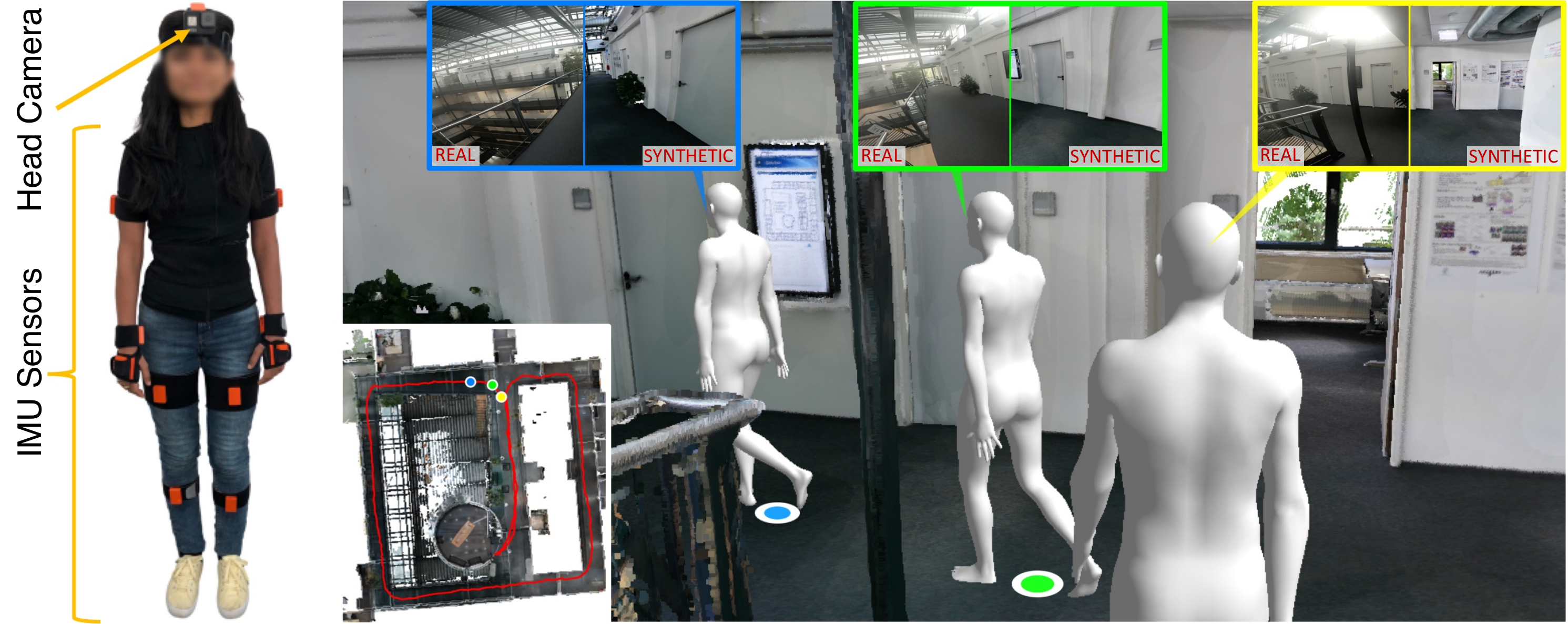}
	\end{center}
    \refstepcounter{figure}\normalfont Figure~\thefigure. HPS jointly estimates the full 3D human pose and location of a subject within large 3D scenes, using only wearable sensors. Left: subject wearing IMUs and a head mounted camera. Right: using the camera, HPS localizes the human in a pre-built map of the scene (bottom left). The top row shows the split images of the real and estimated virtual camera.
	\label{fig:teaser}
	\newline
}
\makeatother

\maketitle
\thispagestyle{empty}
\enlargethispage{\footskip}
\begin{abstract}
\blfootnote{* Joint first authors with equal contribution.}
We introduce (HPS) Human POSEitioning System, a method to recover the full 3D pose of a human registered with a 3D scan of the surrounding environment using wearable sensors. 
Using IMUs attached at the body limbs and a head mounted camera looking outwards, HPS fuses camera based self-localization with IMU-based human body tracking.
The former provides drift-free but noisy position and orientation estimates while the latter is accurate in the short-term but subject to drift over longer periods of time.

We show that our optimization-based integration exploits the benefits of the two, resulting in pose accuracy free of drift. 
Furthermore, we integrate 3D scene constraints into our optimization, such as foot contact with the ground, resulting in physically plausible motion. 
HPS complements more common third-person-based 3D pose estimation methods. It allows capturing larger recording volumes and longer periods of motion, and could be used for VR/AR applications where humans interact with the scene without requiring direct line of sight with an external camera, or to train agents that navigate and interact with the environment based on first-person visual input, like real humans. 

With HPS, we recorded a dataset of humans interacting with large 3D scenes (300-1000 $m^2$) consisting of 7 subjects and more than 3 hours of diverse motion. The dataset, code and video will be available on the project page: \href{http://virtualhumans.mpi-inf.mpg.de/hps/}{http://virtualhumans.mpi-inf.mpg.de/hps/}. 
\end{abstract}
\vspace{-4mm}
\section{Introduction}
\label{sec:introduction}
\input{sections/introduction}

\section{Related Work}
\label{sec:related}
\input{sections/related}

\section{Method}
\label{sec:method}
\input{sections/method}

\section{Dataset}
\label{sec:dataset}
\input{sections/dataset}

\section{Experiments}
\label{sec:experiments}
\input{sections/experiments}

\section{Conclusions and Future Work}
\label{sec:conclusion}
\input{sections/conclusions}

\footnotesize
\noindent
\textbf{Acknowledgments: }We thank Bharat Bhatnagar, Verica Lazova, Anna Kukleva and Garvita Tiwari for their feedback. This  work  is  partly  funded  by  the DFG - 409792180 (Emmy Noether Programme, project:  Real Virtual Humans), 
the EU Horizon 2020 project RICAIP (grant agreeement No.857306), and the European Regional Development Fund under  project IMPACT (No.~CZ.02.1.01/0.0/0.0/15 003/0000468).
{\small
\bibliographystyle{ieee_fullname}
\bibliography{egbib}
}

\end{document}

%% file: notation.tex
\renewcommand{\vec}[1]{\boldsymbol{#1}}
\newcommand{\mat}[1]{\mathbf{#1}}

\newcommand{\real}[0]{\mathbb{R}}

\newcommand{\pose}[0]{\vec{\theta}}
\newcommand{\shape}[0]{\vec{\beta}}

\newcommand{\trans}[0]{\vec{t}}

\newcommand\reallywidehat[1]{%
\savestack{\tmpbox}{\stretchto{%
  \scaleto{%
    \scalerel*[\widthof{\ensuremath{#1}}]{\kern-.6pt\bigwedge\kern-.6pt}%
    {\rule[-\textheight/2]{1ex}{\textheight}}
  }{\textheight}%
}{0.5ex}}%
\stackon[1pt]{#1}{\tmpbox}%
}

%% file: sections/introduction.tex
Capturing the full 3D pose of a human, while localizing and registering it with a 3D reconstruction of the environment, using \emph{only wearable sensors}, opens the door to many applications and new research directions. 
For example, it will allow Augmented / Mixed / Virtual Reality users to move freely and interact with virtual objects in the scene, without the need for external cameras.
From the captured data, we could train digital humans that plan and move like real humans, based on visual data arriving at their eyes.
Moreover, by relying only on ego-centric data, we could capture a wider variety of human motion, outside of a restricted recording volume imposed by external cameras. 

The dominant approach in vision has been to analyze humans from an \emph{external third-person camera}, often without considering scene context~\cite{PonsModelBased,kanazawa2018endtoend,omran2018neural,alldieck2019tex2shape,luo20203d,sarandi2020metrabs}. A few recent methods capture 3D scenes and humans~\cite{hassan2019prox}, but again using a third-person camera. 
Capturing with external cameras is undoubtedly a central problem in vision, but it has its limitations -- occlusions are a problem, and interactions across multiple rooms or beyond the viewing area cannot be captured; consequently recordings are typically short. 

We propose \emph{Human POSEitioning System} (HPS), the first method to recover the full body 3D pose of a human registered with a large 3D scan of the surrounding environment relying \emph{only on wearable sensors} --  body-mounted IMUs and a head mounted camera, approximating the visual field of view of the human. Inspired by visual-inertial odometry and localization~\cite{Lynen2015RSS,Jones2011IJRR}, as well as IMU-based human pose estimation~\cite{pons2010multisensor,vonMarcard2018,SIP}, HPS fuses information coming from body-mounted IMUs with camera pose obtained from camera self-localization~\cite{sattler2016efficient,sarlin2019coarse,Taira2018CVPR} (see Fig.~\ref{fig:teaser}). Instead of placing the camera towards the body~\cite{rhodin2016egocap,SelfPose2020}, we place it towards the scene, which allows us to capture what the human observes together with their 3D pose. In comparison to third-person pose methods, the body is not seen by the camera, which poses new challenges. 

Pure IMU-based tracking is known to drift over time and camera localization produces many outliers. By jointly integrating IMU tracking with camera self-localization, we are able to remove drift~\cite{Lynen2015RSS,Jones2011IJRR}, and recover the human trajectory when self-localization fails. 
Furthermore, since we can approximately locate the person in the 3D scene, we incorporate scene constraints when foot contact is detected. Overall, with HPS we recover natural human motions, registered with the 3D scene and free of drift, during \emph{long periods} of time, and over \emph{large areas}.

To demonstrate the capabilities of HPS, we capture a dataset of real people moving in large scenes. 
Our HPS dataset consists of $8$ types of environments - some being larger than $1000 m^2$, and $7$ subjects performing a variety of activities such as walking, excercising, reading, eating, or simply working in the office.
The dataset can be used as a testbed for ego-centric tracking with scene constraints, to learn how humans interact and move within large scenes over long periods of time, and to learn how humans process visual input arriving at their eyes. 

We make the following contributions: 
\textbf{1}) to the best of our knowledge, HPS is the first approach to estimate the full 3D human pose while localizing the person within a pre-scanned large 3D scene using wearable sensors. 
\textbf{2}) we introduce a joint optimization which integrates camera localization, IMU-based tracking and scene constraints, resulting in smooth and accurate human motion estimates. 
\textbf{3}) we provide the \emph{HPS dataset}, a new dataset consisting of 3D scans of large scenes (some larger than 1000 $m^2$), ego-centric video, IMU data, and our 3D reconstructed humans moving and interacting with the scene. 
In contrast to existing 3D pose datasets, which are captured from a third-person view, ours is captured from an egocentric view. 
We believe both HPS and HPS dataset will provide a step towards developing future algorithms to understand and model 3D human motion and behavior within the 3D environment from an egocentric (or third-person) perspective.

%% file: sections/related.tex
\begin{figure*}
    \centering
    \includegraphics[width=\textwidth]{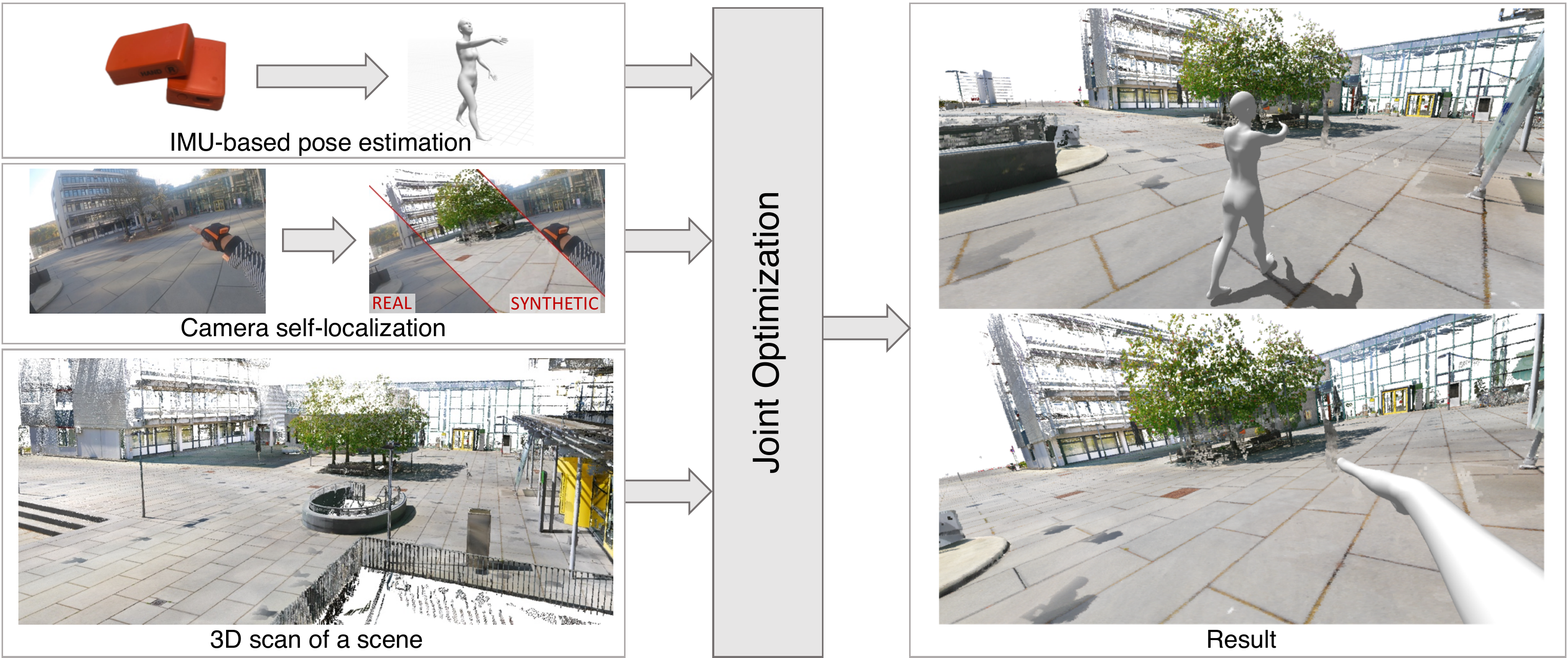}
    \caption{\textbf{Overview.} We use IMU data, RGB video from a head mounted camera, and a pre-scanned scene  as input. We obtain an  approximate 3D body pose using IMU data, and use head camera self-localization to localize the subject in the 3D scene. 
    We then integrate the approximate body pose, the camera position and orientation, along with the 3D scene in a joint optimization to obtain the final location and pose estimates. We urge readers to see the video at \href{http://virtualhumans.mpi-inf.mpg.de/hps/}{http://virtualhumans.mpi-inf.mpg.de/hps/}. }
    \label{fig:overview}
    \vspace{-4mm}
\end{figure*} 
\textbf{IMU-based 3D Human Pose Estimation: }Although commercial solutions for IMU-based pose estimation have improved the stability of earlier solutions~\cite{roetenberg2007moven}, they still suffer from severe drift, especially in the global orientation and location of the body. 
Early work~\cite{vlasic2007practical} developed a custom suit to capture 3D human pose during daily activities. 
One line of work has focused on reducing the amount of IMUs necessary to capture motion via space-time optimization~\cite{SIP} or with deep learning~\cite{DIP:SIGGRAPHAsia:2018}. 
In order to reduce drift and improve accuracy, visual-inertial approaches combine IMUs with multiple external cameras~\cite{Marcard2016,pons2010multisensor,Pons-Moll2011,TrumbleBMVC17,Malleson3DV17}, a depth-camera ~\cite{Helten:2013,zherong2018} or even a single hand-held RGB camera~\cite{vonMarcard2018}--which allowed collecting the 3DPW~\cite{vonMarcard2018} dataset with accurate 3D poses outdoors. However, they all require an external camera, which limits the field of view to be captured, or requires someone to follow the person being tracked. Instead, we mount the camera (approximating the person's field of view) on the head and use it to self-localize the person in the scene. 

\textbf{Ego-centric capture and prediction: }In contrast to our method, most ego-centric body-capture approaches mount the camera on the head looking towards the body.
While ego-centric capture has received considerable attention for activity recognition~\cite{ijcai2017-200,fathi2011understanding,ma2016going,cao2017egocentric,yonemoto2015egocentric,rogez2015first}, methods at most detect the upper body. 
For full body capture, a pioneering method ~\cite{rhodin2016egocap} relied on a helmet with sticks holding a camera away from the body.
More recent methods~\cite{xu2019mo2cap2,SelfPose2020} work reasonably well even when the camera is close to the head.
However, the accuracy is still far from desired.  

Another group of methods place the camera looking outwards (like humans), and aim at estimating 3D pose from the ego-centric view alone, but 3D poses are inaccurate and have high uncertainty~\cite{jiang2017seeing,yuan20183d,Yuan_2019_ICCV}.
These methods to infer 3D pose from an ego-centric view~\cite{jiang2017seeing,yuan20183d,Yuan_2019_ICCV} would benefit from our captured data, which contains ego-centric video with corresponding accurate 3D pose registered with the environment.
An alternative approach places many cameras on the body looking out and use multi-camera structure from motion~\cite{shiratori2011motion}, but it can only recover slow motions. 

\textbf{Camera Localization: }Most 6-DoF camera localization algorithms can be split into three groups. 
The first group is \emph{structure-based}~\cite{svarm2016city,toft2018semantic,sattler2016efficient,liu2017efficient,taira2018inloc,Shotton2013CVPR,brachmann2020ARXIV,Cavallari2019TPAMI}, which matches 2D points in the query image with 3D scene keypoints to estimate the camera pose by minimizing the reprojection error. 
While they provide precise position in small scenes, they do not scale to large scenes as matching becomes ambiguous and computationally expensive.

The second group of methods is referred to as \emph{image-based}. The idea is to retrieve nearest neighbors in an image database 
based on a global descriptor ~\cite{arandjelovic2016netvlad,torii201524,weyand2016planet}. 
The camera pose can then be approximated by the known poses of the retrieved images. 
They are more robust and scalable compared to structure based methods, but less precise, and the quality  depends on the size of the image database. 

In the third group are \emph{hybrid approaches}~\cite{sarlin2019coarse, sarlin2018corl, Brachmann2019ICCVa} which combine the benefits of the last two. First, a set of relevant database images are found using an image-based method, and then the precise camera pose is recovered using structure-based methods.  Another set of methods directly regress the camera pose using a CNN~\cite{Sattler2019CVPR,Walch2017ICCV}, but their accuracy leaves a lot to be desired. \emph{Hybrid approaches} have been shown to be precise and to scale to large scenes, and hence the self-localization part of HPS builds upon them. 

\textbf{Humans and Scenes:}
The relationship between humans, scenes, and objects is a recurrent subject of study in vision.
Examples are methods for 2D pose and object detection~\cite{desai2012detecting,yao2010modeling,iqbal2017pose,pishchulin2013poselet,kjellstrom2011visual,gupta2007objects}, 3D object detection using human poses~\cite{grabner2011makes,gupta20113d}, learning to insert people in scenes~\cite{fouhey2014people,wang2017binge,li2019putting,zhang2020generating}, constraining pose~\cite{zanfir2018monocular,hassan2019prox}, estimating forces~\cite{li2019estimating}, or predicting long term motion \cite{caoHMP2020} conditioned on the scene.
\emph{Most approaches predict only static poses in a single room}, and reasoning is done from a third-person perspective. In contrast, our analysis is from a first-person perspective, and uses the scene to self-localize the human in it. Furthermore, our method enables to capture humans in motion in multiple-room and outdoor environments. All aforementioned methods would benefit from the HPS dataset.

%% file: sections/method.tex
Our goal is to recover the 3D body pose and location of a subject in a known scene from egocentric measurements. To this end, our method requires as input: $1)$ a head-mounted camera,  $2)$ body-mounted IMUs, and $3)$ a pre-built 3D scan of a scene, along with a database of RGB scene images with known camera parameters.
Using camera data, our method localizes the person within a pre-scanned 3D scene  (Sec.~\ref{subsec:meth_cam_loc}), estimates their 3D pose using IMUs (Sec.~\ref{subsec:meth_pose_est}), and in a joint optimization step  (Sec.~\ref{subsec:meth_sens_fusion}) integrates camera localization, IMU pose estimates and scene  constraints, resulting in smooth and accurate human motion estimates. For an overview of our method, see Fig.~\ref{fig:overview}. 
For more details on the 3D scene reconstruction, 
image database collection, 
camera and IMU setup, we refer to the supplementary. 
\subsection{SMPL Body Model}
\label{subsec:meth_smpl}
We use the Skinned Multi-Person Linear (SMPL) body model  \cite{smpl2015loper} to represent the human subject. SMPL is a differentiable function $M(\pose, \trans, \shape) : \real^{72 \times 3 \times 10} \mapsto \real^{6890 \times 3}$ that maps pose $\pose$, translation $\trans$ and shape $\shape$ parameters to the vertices of a watertight human mesh.
The underlying skeleton of SMPL has $24$ joints. The pose parameters $\pose \in \real^{72} $ correspond to the relative orientation of each joint in the SMPL skeleton, expressed in axis-angles. The shape parameters $\shape \in \real^{10}$ are the PCA coefficients of a shape space learnt from a corpus of registered scans. We use the notation $M_n(\pose, \trans, \shape) \in \real^3$ to indicate the $n^{th}$ vertex of SMPL. We obtain approximate shape parameters $\shape$ of a person from body measurements. We assume that $\shape$ remains constant during a sequence and aim to recover $\pose$ and $\trans$ of the subject registered with the 3D environment. 
Henceforth we drop $\shape$ for notational convenience. 
\begin{figure}
    \centering
    \includegraphics[width=\columnwidth]{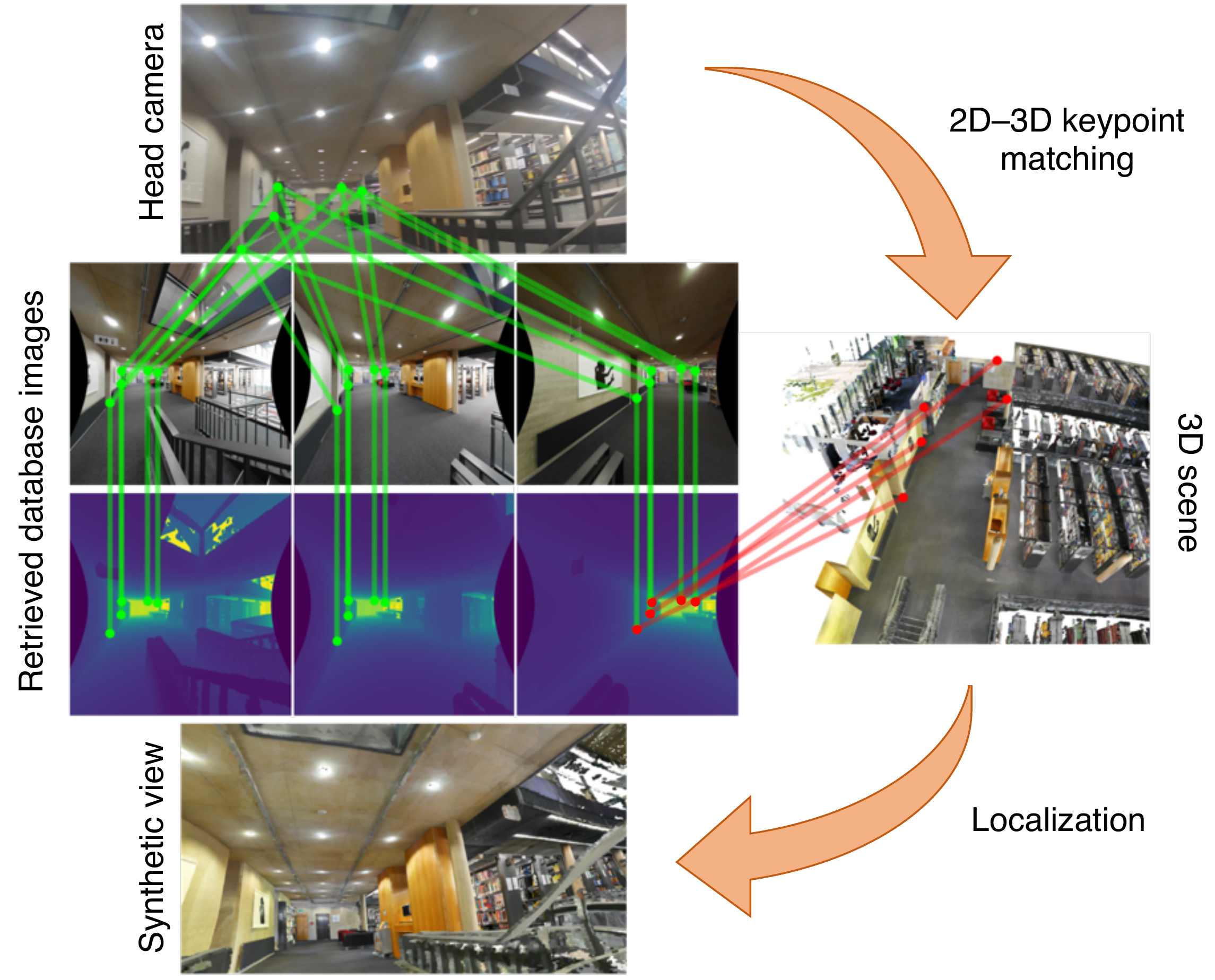}
    \caption{\textbf{Camera self-localization.} We match the head camera image keypoints with the keypoints from the prefiltered database with known 2D-3D scene correspondences. We then localize the camera in the scene by minimizing a reprojection error of the keypoints. \emph{From top to bottom:} head camera image (query), top-3 retrieved images from a dataset, depthmaps rendered from the same position to map 2D database keypoints to 3D, synthetic view of the scene from the inferred camera position.}
    \vspace{-2mm}
    \label{fig:meth_cameraloc_pipeline}
\end{figure}
\subsection{Camera Self-localization}
\label{subsec:meth_cam_loc}
The camera self-localization stage aims to estimate the position and orientation of the human head from a head-mounted camera. 
To scale to large scenes, we use a hierarchical structure-based localization algorithm~\cite{sarlin2019coarse,sarlin2020superglue} (Fig.~\ref{fig:meth_cameraloc_pipeline}). 
It first identifies a set of potentially relevant database images, \ie, images used to build the 3D scene map, through image retrieval via NetVLAD~\cite{arandjelovic2016netvlad} descriptors. 
2D-3D matches are established between local SuperPoint~\cite{detone2018superpoint} features extracted in the query image and 3D points visible in the top-40 retrieved images. 
These matches are then used to estimate the camera pose by applying a P3P solver~\cite{Kneip2011CVPR,Haralick1994IJCV,Kukelova2010ACCV} inside a RANSAC loop~\cite{Fischler81CACM} with local optimization~\cite{Lebeda2012BMVC}. 
Rather than building a separate sparse Structure-from-Motion point cloud for localization, as originally used in~\cite{sarlin2019coarse}, we obtain 3D point positions from our dense scene 3D model~\cite{Taira2018CVPR}. 
For each pixel in a database image, we obtain the corresponding 3D point by rendering the 3D model from the known pose of the image. 
2D-2D matches between the query and the top-40 retrieved database images thus yield the required 2D-3D matches.
From the camera self-localization step, we obtain estimates for camera orientation $\mat{R}^C$ and position $\vec{t}^C$.
\subsection{IMU based Pose Estimation}
\label{subsec:meth_pose_est}
\begin{figure}
    \centering
    \includegraphics[width=.8\columnwidth]{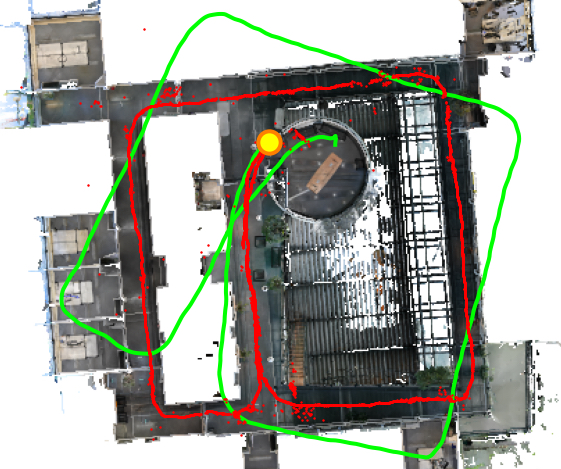}
    \caption{\textbf{Comparison of the trajectories} of IMUs (in green) with camera self-localization (in red). The yellow dot marks the start. Notice the red trajectory is free of drift but has outliers.}
    \vspace{-2mm}
    \label{fig:exp_imu_camera_drift}
\end{figure}

We use a commercial inertial mo-cap system provided by XSens \cite{xsensawinda}, which uses 17 IMUs attached to the body with velcro-straps or a suit. 
XSens IMUs provide 3D pose estimates, denoted as $\pose^I$ and location estimates relative to the starting position of a recording - denoted as $\vec{t}^I$, using a proprietary algorithm based on a Kalman filter and a kinematic model of the human body to reduce drift. 
While it provides accurate articulation, our experiments show that the global orientation and position drift significantly over time, and consequently scene constraints are not satisfied (Fig.~\ref{fig:exp_imu_camera_drift}, \ref{fig:optim_improvement_qualitativ}). 
Using acceleration information, IMUs also detect feet contacts with the ground, which we integrate in our joint optimization algorithm. 
\subsection{Joint Optimization}
\label{subsec:meth_sens_fusion}
Our joint optimization algorithm finds the pose parameters of the SMPL body model in order to satisfy i) the head camera self-localization, ii) scene, and iii) smoothness constraints while remaining as close as possible to the IMU pose estimate $\pose^I$ (excluding global orientation and position) -- while we could optimize SMPL to match the raw IMU data directly~\cite{vonMarcard2018,SIP}, we chose not to, because it contains a lot of drift.
Mathematically, we minimize
the following objective over a batch of $T$ frames ($T$ is fixed for all scenes)
\begin{equation}
    E(\pose_{1:T}, \trans_{1:T}) = w_s E_\mathrm{self} + w_{sc} E_\mathrm{scene}  + w_\mathrm{sm}E_\mathrm{sm} + w_p E_\mathrm{IMU},
    \label{eq:objective}
\end{equation}
with respect to pose $\pose_{1:T}$ and translation $\trans_{1:T}$ parameters. $\pose_{1:T} \in \real^{72T}$ and $\vec{t}_{1:T} \in \real^{3T}$ are stacked model poses and translations for each time step $j = 1 \dots T$. In the following, we explain each of the terms in more detail.

\PAR{Self-localization Term $E_\mathrm{self}$:} 
We use the estimated orientation of the camera to constrain the orientation of SMPL.
Specifically, we minimize the geodesic distance~\cite{SIP} from the head camera orientation as inferred from SMPL, $\overline{R}^{C}(\pose)$, to the self-localization estimate $\mat{R}^{C}$ over a batch of frames $T$: 
\begin{equation}
E_\mathrm{self} =\frac{1}{T}\sum_{j = 1}^{T}||(\mathrm{log}((\overline{R}^{C}(\pose_j))^\top \mat{R}^{C}_{j}))^{\vee}||_2 \enspace,
\vspace{-2mm}
\end{equation}
where the $\mathrm{log}$ operation recovers the skew-symmetric matrix from the relative rotation matrix, and the $^{\vee}$ operator converts it to its axis-angle representation. 
The mapping $\overline{R}^{C}(\pose_j)$ can be derived as follows. First, we obtain the head bone orientation by traversing the kinematic chain of SMPL
\begin{equation}
R^{H} (\pose)= {\displaystyle \prod_{i  \in \mathcal{P}_\mathrm{Head}}} \mathrm{exp}(\reallywidehat{\pose^i}) \enspace,
\label{eq:head}
\vspace{-2mm}
\end{equation}
where $\mathcal{P}_\mathrm{Head}$ is an ordered list of all the parents to the head joint. The $^{\wedge}$ operator maps an axis-angle to its corresponding skew-symmetric matrix and  $\mathrm{exp}(\reallywidehat{\pose^i})$ are the relative joint rotation matrices obtained from $\pose^i\in so(3)$ using the Rodrigues formula.
While $R^H : \real^{72} \mapsto \mathrm{SO(3)}$ maps from pose to head rotation, we need a mapping to camera orientation. 
Since the camera is rigidly attached to the head, there is a constant camera to head offset that can be estimated at frame $0$~\cite{SIP,pons2010multisensor}:
\begin{equation}
\mat{R}_{HC} = (R^{H}(\pose^I_0 ))^\top\mat{R}^{C}_{0} \enspace.
\end{equation}
We find the desired mapping from pose to camera at a subsequent frame $j$ as $\overline{R}^{C}(\pose_j) = R^{H}(\pose_j)\ \mat{R}_{HC}$.
\PAR{Scene Contact Term $E_\mathrm{scene}$:}
When the IMUs detect a foot contact, we force it to be in contact with the ground by using an energy term consisting of two subterms $E_\mathrm{scene} = w_c E_\mathrm{contact} + w_v E_\mathrm{slide}$.
Let $\mathcal{B}_k$ with $k \in [1,2,3,4]$ denote 4 sets of manually defined vertex indices in the SMPL corresponding to the toe and heel regions for the left and right foot (more details in supplementary), and let $c^k_j \in [0,1]$ be a binary variable indicating if part $k$ is in contact with the ground at frame $j$. 
We define the following contact term, which snaps the foot vertices to the closest scene vertices
\begin{equation}
\small
E_\mathrm{contact} = \frac{1}{4T}\sum_{j = 1}^{T}\sum_{k = 1}^{4}\sum_{n \in \mathcal{B}_k} \frac{1}{|\mathcal{B}_k|} c^{k}_{j}||M_n(\pose_j, \vec{t}_j) - v(n)\|_2 \enspace,
\end{equation}
where $M_n(\pose_j, \trans_j)$ is the $n^{th}$ vertex of the SMPL mesh at frame $j$, and $v(n)=\underset{\vec{v}_s \in \mat{V_s}}{\mathrm{argmin}}(||M_n(\pose_j, \vec{t}_j) - \vec{v}_s||_2)$ returns the closest scene point $\vec{v}_s \in \mat{V_s}$ to $M_n(\pose_j, \trans_j)$.
To prevent the foot from sliding when in contact with the scene, we also constrain the distance between foot parts in contact with the scene in two successive frames to be zero.
\begin{multline}
    E_\mathrm{slide} = \frac{1}{4(T-1)}\sum_{j = 1}^{T-1}\sum_{k = 1}^{4} \sum_{n \in \mathcal{B}_k} \frac{1}{|\mathcal{B}_k|} c_{j}^{k}c_{j+1}^k||M_n(\pose_j, \vec{t}_j) - \\ 
    M_n(\pose_{j+1}, \vec{t}_{j+1}) ||_2 \enspace.
    \vspace{-1.5mm}
\end{multline}
\PAR{Smoothness Term $E_{\mathrm{sm}}$: } This term ensures smooth changing of the global translation and orientation, as well as head orientation
\vspace{-3mm}
\begin{equation}
    E_\mathrm{sm} =  w_{T}E_{T} + w_{G}E_{G} + w_{H}E_{H},
\end{equation}
where the translation term equals:
\begin{equation}
    E_{T} = \frac{1}{T - 1}\sum_{j = 1}^{T - 1}||(\vec{t}_j - \vec{t}_{j + 1})||_2 \enspace.
    \vspace{-2.5mm}
\end{equation}
Defining $R^G : \real^{72} \mapsto \mathrm{SO(3)}$ as $R^G (\pose) = \mathrm{exp}(\reallywidehat{\pose^{G}})$ where $\pose^{G}$ is the axis-angle representation of the root (global) joint, the global orientation smoothness term is 
\begin{equation}
    E_{G} =\frac{1}{T - 1}\sum_{j = 1}^{T - 1}||(\mathrm{log}((R^{G}(\pose_{j}))^{\top} R^{G}(\pose_{j + 1})))^{\vee}||_2 
    \label{eq:smoothness_glob}
\end{equation}
Using Eq~\eqref{eq:smoothness_glob}, the head orientation smoothness term is enforced with an equivalent term replacing $R^G$ by $R^H$.
\PAR{Pose Term $E_\mathrm{IMU}$:} The pose recovered by IMUs captures the articulation of the body well, but is inaccurate for global orientation and translation. 
Hence, we constrain the pose parameters corresponding to the body to remain close to the IMUs estimate. Let $\mat{B}$ be an identity matrix with zeros at the diagonal entries corresponding to the root joint. With this, the pose is regularized with the following equation:
\begin{equation}
    E_\mathrm{IMU} =\frac{1}{T}\sum_{j = 1}^{T} \sqrt{(\pose_j - \pose_j^I)^\top \mat{B} (\pose_j - \pose_j^I)} \enspace.
    \vspace{-1mm}
\end{equation}
For implementation details of our joint optimization algorithm, please see the supplementary.
\subsection{Initialization}
\label{subsec:meth_initialization}
Since the objective function in Eq.~\eqref{eq:objective} is highly non-convex, convergence to a good minimum hinges on good initialization. 
We initialize translation parameters $\trans_j$ using camera localization estimates $\vec{t}^C_j$.
Camera localization results are typically noisy, so instead of using raw results we first detect outliers by computing the velocity of translation between each result and its inlier neighbours. We mark a result as an outlier if its velocity exceeds the threshold $\epsilon=3m/s$. We repeat this process until convergence and replace all outliers by interpolation. \par
For  poses $\pose$, the simplest choice is to initialize with the IMU pose estimate $\pose_j = \pose^I_j$. 
However, the global body orientation often deviates from the more accurate self-localization trajectory (see Fig.~\ref{fig:exp_imu_camera_drift}, \ref{fig:optim_improvement_qualitativ}). 
Observing that the body orientation is often perpendicular to the trajectory, our idea is to rotate the IMU pose to align it to the self-localization trajectory. To this end, we first estimate the tangent direction of the self-localization and IMU trajectories
\vspace{-3mm}
$$\vec{v}^C_{j} = \frac{\vec{t}^{C}_{j+\gamma} - \vec{t}^{C}_{j}}{||\vec{t}^{C}_{j+\gamma} - \vec{t}^{C}_{j}||_2 } \enspace , \quad \vec{v}^I_{j} = \frac{\vec{t}^{I}_{j+\gamma} - \vec{t}^{I}_{j}}{||\vec{t}^{I}_{j+\gamma} - \vec{t}^{I}_{j}||_2} \enspace,$$
($\gamma = 10$ in our case) and correct the root orientation $\mathrm{exp}(\reallywidehat{\pose^{I,G}_j})$ of the IMU pose with the following formula
\vspace{-2mm}
\begin{equation}
\pose^{I,G*}_j = (\mathrm{log}(\mathrm{exp}(\reallywidehat{\vec{v}^{I}_{j} \times \vec{v}^C_{j}})\mathrm{exp}(\reallywidehat{\pose^{I,G}_j})))^{\vee} \enspace,
\end{equation}
\vspace{-1mm}
where $\mathrm{exp}(\reallywidehat{\vec{v}^{I}_{j} \times \vec{v}^C_{j}})$ is the planar rotation that aligns $\vec{v}^I_{j}$ with $\vec{v}^C_{j}$.
For stationary frames, we use the correction matrix of the last frame with non-zero velocity. We find that in practise, for stationary frames, this a good approximation.
\subsection{Coordinate Frame Alignment}
\label{subsec:meth_axis_align}
While the camera estimates $\mat{R}^C$ and $\vec{t}^C$ are in the 3D scene coordinates, IMU estimates $\pose^{I}$ and $\vec{t}^{I}$ are not. Before the initialization step (Sec.~\ref{subsec:meth_initialization}) of our joint optimization algorithm  (Sec.~\ref{subsec:meth_sens_fusion}), we align the IMU coordinate frame with the 3D scene frame by finding a planar rotation $\mat{R}^{*}_A$ that orients the SMPL head  at frame zero $R^H(\pose^I_0)$ to match the camera orientation $\mat{R}^{C}_{0}$ at the same frame. Mathematically, this entails minimizing the following objective
\begin{equation}
    \mat{R}^{*}_A = \underset{\mat{R}_A \in \mathcal{R}}{\mathrm{argmin}}||(\mathrm{log}(\mat{R}_A R^H(\pose^I_0))^{\top}\mat{R}^{C}_{0}))^{\vee}||_2 \enspace .
\end{equation}

We use the axis-angle parameterization to define the set of rotation matrices $\mathcal{R} = \{ \mathrm{exp}(\reallywidehat{x\vec{\alpha}}) : x \in \real \}$. 
where $\vec{\alpha} = [0, 0, 1]^\top$ is the z-axis unit vector. The IMU pose $\pose^I_j$ and position $\vec{t}^{I}_j$  estimate of each subsequent frame are aligned to the 3D scene reference frame by
\vspace{-2mm}
\begin{equation}
\pose^{I,G}_j = (\mathrm{log}(\mat{R}^{*}_A\mathrm{exp}(\reallywidehat{\pose^{I,G}_j})))^{\vee} \enspace , 
\vec{t}^{I}_j = \mat{R}^{*}_A \vec{t}^{I}_j\enspace .
\end{equation}

%% file: sections/dataset.tex
\emph{HPS} allows us to collect the \emph{HPS dataset} - a dataset of 3D humans interacting with large 3D scenes (300-1000~$m^2$, up to 2500~$m^2$). Our dataset contains images captured from a head-mounted camera coupled with the reference 3D pose and location of the person in a pre-scanned 3D scene. We capture 7 people in 8 large scenes performing
activities such as exercising, reading, eating, lecturing, using a computer, making coffee, dancing. \emph{All subjects have agreed to release their data for research purposes.} 
In total, the dataset provides more than 300K synchronized RGB images coupled with the reference 3D pose and location. We plan to keep updating the dataset by adding more long-term motion recordings with a variety of scene interactions. Figure \ref{fig:exp_qulitative_results} shows qualitative results from our dataset. For more examples, please see the video~\cite{hps_project_page}.

%% file: sections/experiments.tex
This section shows that HPS does not drift with time and distance traveled, is robust to non-persistent camera localization outliers, and satisfies scene constraints (feet stay on the ground during contact, and do not slide). 

Since this is the first method to track humans in large scenes, there exist no published baselines to compare to, and ground truth 3D human pose and localization cannot be obtained for unbounded areas like ours. Hence, we use depth cameras to obtain ground truth dynamic point clouds of the human in a small sub-area of the scene. Subjects are then asked to move freely in the large scene, and return to the sub-area, where we can evaluate accuracy and drift. 
 
\begin{table}
\begin{center}
    \footnotesize
	\tabcolsep=0.11cm
	\begin{tabular}{|l||c|c|c|c|c|} \hline 
		\begin{tabular}{@{}l@{}}Distance\\traveled\end{tabular} & 
		IMU & 
		IMU~+~Cam &
		\begin{tabular}{@{}c@{}}IMU~+~Cam\\(filtered)\end{tabular} & \begin{tabular}{@{}c@{}}HPS\\w\textbackslash o scene\end{tabular} & HPS      \\ \hline \hline 
At start & 6.85 & 9.24 & 10.48 & 7.21 & \textbf{5.20} \\ \hline
70 m & 54.49 & 742.32 & 6.93 & 6.48 & \textbf{4.60} \\ \hline
200 m & 69.02 & 136.81 & 5.93 & 5.80 & \textbf{4.26} \\ \hline
380 m & 108.44 & 32.17 & 6.15 & 5.69 & \textbf{4.53} \\ \hline

	\end{tabular}
\end{center}
\vspace{-1.5mm}
\caption{\textbf{Drift and cam. outliers:} 3D error (in cm) for the subject standing in A-pose after moving freely around the scene.}
\label{tab:eval_table_apose}

\begin{center}
    \footnotesize
	\tabcolsep=0.11cm
	\begin{tabular}{|l||c|c|c|c|c|} \hline
		\begin{tabular}{@{}l@{}}Distance\\traveled\end{tabular} & 
		IMU & 
		IMU~+~Cam &
		\begin{tabular}{@{}c@{}}IMU~+~Cam\\(filtered)\end{tabular} & \begin{tabular}{@{}c@{}}HPS\\w\textbackslash o scene\end{tabular} & HPS      \\ \hline \hline 
At start & 6.77 & 2189.75 & 10.05 & 9.19 & \textbf{6.44} \\ \hline
70 m & 51.57 & 569.71 & 21.75 & 20.68 & \textbf{15.96} \\ \hline
200 m & 61.11 & 719.44 & 7.34 & 6.67 & \textbf{4.76} \\ \hline
380 m & 100.44 & 261.72 & 12.59 & 11.96 & \textbf{10.07} \\ \hline

	\end{tabular}
\end{center}
\vspace{-1.5mm}
\caption{\textbf{Drift and cam. outliers (dynamic):} 3D error (in cm) for the subject walking, standing and leaning on the table, after moving around the scene. Error is measured from the dynamic ground truth point cloud to the result (3D mesh in motion). Rows indicate distance traveled before evaluation.}
\label{tab:eval_table_variousact}
\vspace{-3mm}
\end{table}

\subsection{Quantitative Evaluation}
\label{subsec:exp_quantit_eval}
We evaluate the accuracy of our method by comparing our output SMPL mesh (including translation) with a dynamic \emph{ground-truth point cloud} of the person obtained from three   synchronized and calibrated external depth cameras (Azure Kinect\cite{azurekinect}). We register the point cloud to the scene in three steps involving camera self-localization, ICP, and manual correction. For an explanation of the Kinect setup and point cloud registration we refer to the supplementary. 
We report the bidirectional Chamfer distance between the SMPL model (result) and ground truth point cloud from depth sensors \emph{without} Procrustes alignment.
\begin{table}
\begin{center}
    \footnotesize
	\tabcolsep=0.11cm
	\begin{tabular}{|l||c|c|c|c|c|} \hline 
		Metric &  
		IMU &
		IMU~+~Cam &
		\begin{tabular}{@{}c@{}}IMU~+~Cam\\(filtered)\end{tabular} & \begin{tabular}{@{}c@{}}HPS\\w\textbackslash o scene\end{tabular} & HPS     \\ \hline \hline 
	    \begin{tabular}{@{}l@{}}Dist. to Surf.\end{tabular}& 188.38  &   39.8 &     0.95 &   0.32    &    \textbf{0.056} \\ \hline 
		\begin{tabular}{@{}l@{}}Foot Sliding\end{tabular}& 0.92 &      52.09   &   1.75 &    2.00         &  \textbf{0.90}     \\ 

\hline 
	\end{tabular}
\end{center}
\caption{\textbf{Foot contact:} For frames when foot contact is detected, we report (in cm) \textbf{Distance to surface}: Average distance between foot vertices and the scene, and 
\textbf{Foot Sliding}: Average distance on the surface plane between foot vertices in two successive frames.
Numbers are computed for a 3 minute long walking sequence.}
\label{tab:eval_table_feet}
\end{table}

\begin{figure}
    \centering
    \includegraphics[width=\columnwidth]{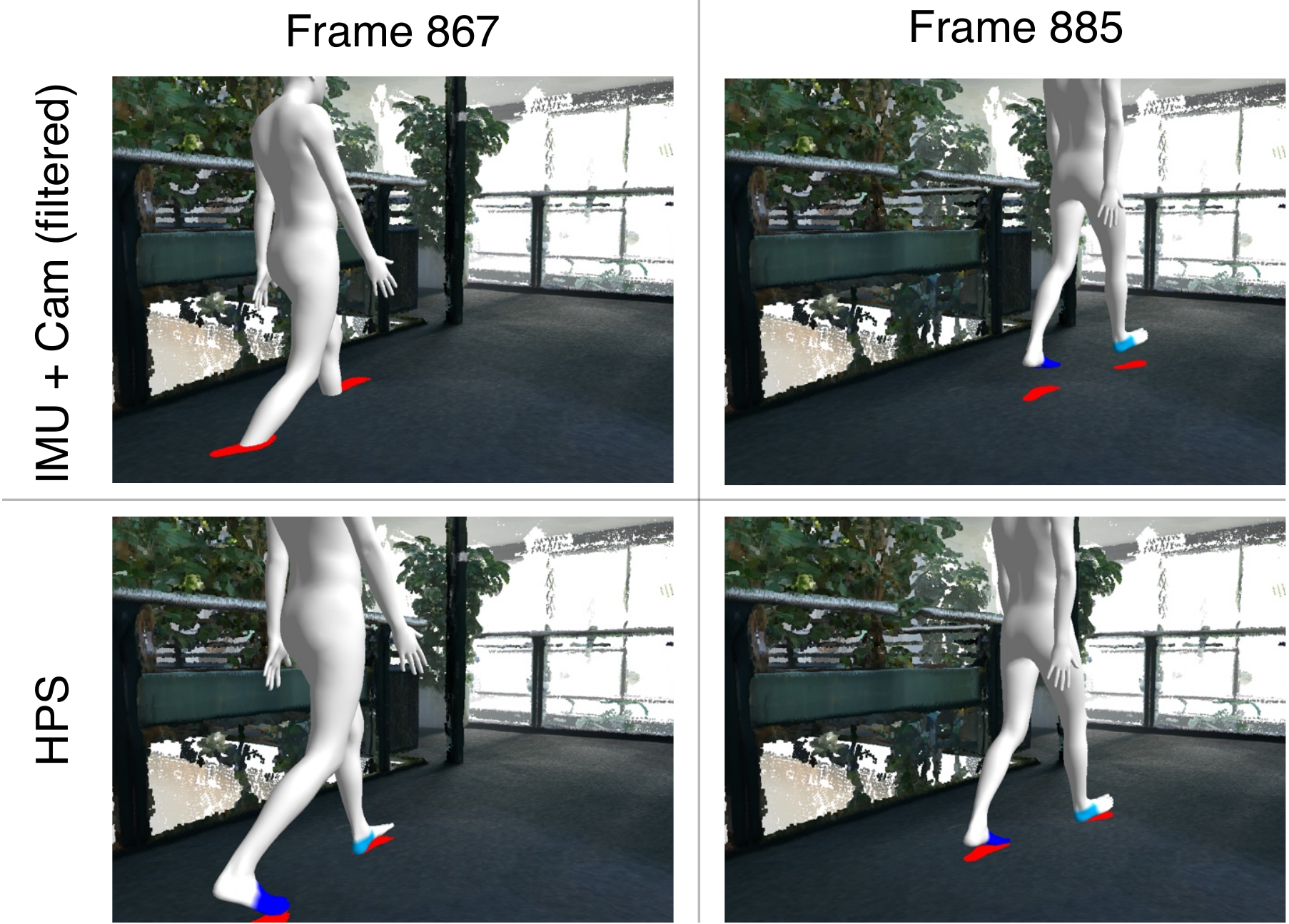}
    \caption{\textbf{Effect of integrating predicted 3D scene contacts.} As a baseline we used camera localization results for localizing SMPL model. Red regions mark closest surface to feet, heels and toes are colored with light blue and blue when IMUs detect ground contact.}
    \vspace{-2mm}
    \label{fig:3D_contacts_qualitativ}
\end{figure}

\textbf{Movements:} For quantitative evaluation, we record using the following protocol: a subject starts within the recording volume of the three RGB-D sensors and performs different actions including standing in A-pose, leaning on a table and walking. The subject then leaves the recording volume and moves within the scene, returns back and repeats the same actions inside that volume again. This is repeated several times, each time choosing a different path. \par
\textbf{Baselines:} There are no established baselines to compare to, as no other method tackles the same problem. 
Hence, to understand the influence of each component, we use 
the following baselines: 1)~\textbf{IMU:} pure IMU tracker, 2)~\textbf{IMU+Cam}: pose from IMU, and translation from camera self-localization, 3)~\textbf{IMU+Cam (filtered)}: Like IMU+Cam but with filtered camera outliers (same as in Sec.~\ref{subsec:meth_initialization}), 4)~\textbf{HPS w\textbackslash o scene}: Optimization without 3D scene contact constraints.

\begin{figure}
    \centering
    \includegraphics[width=\columnwidth]{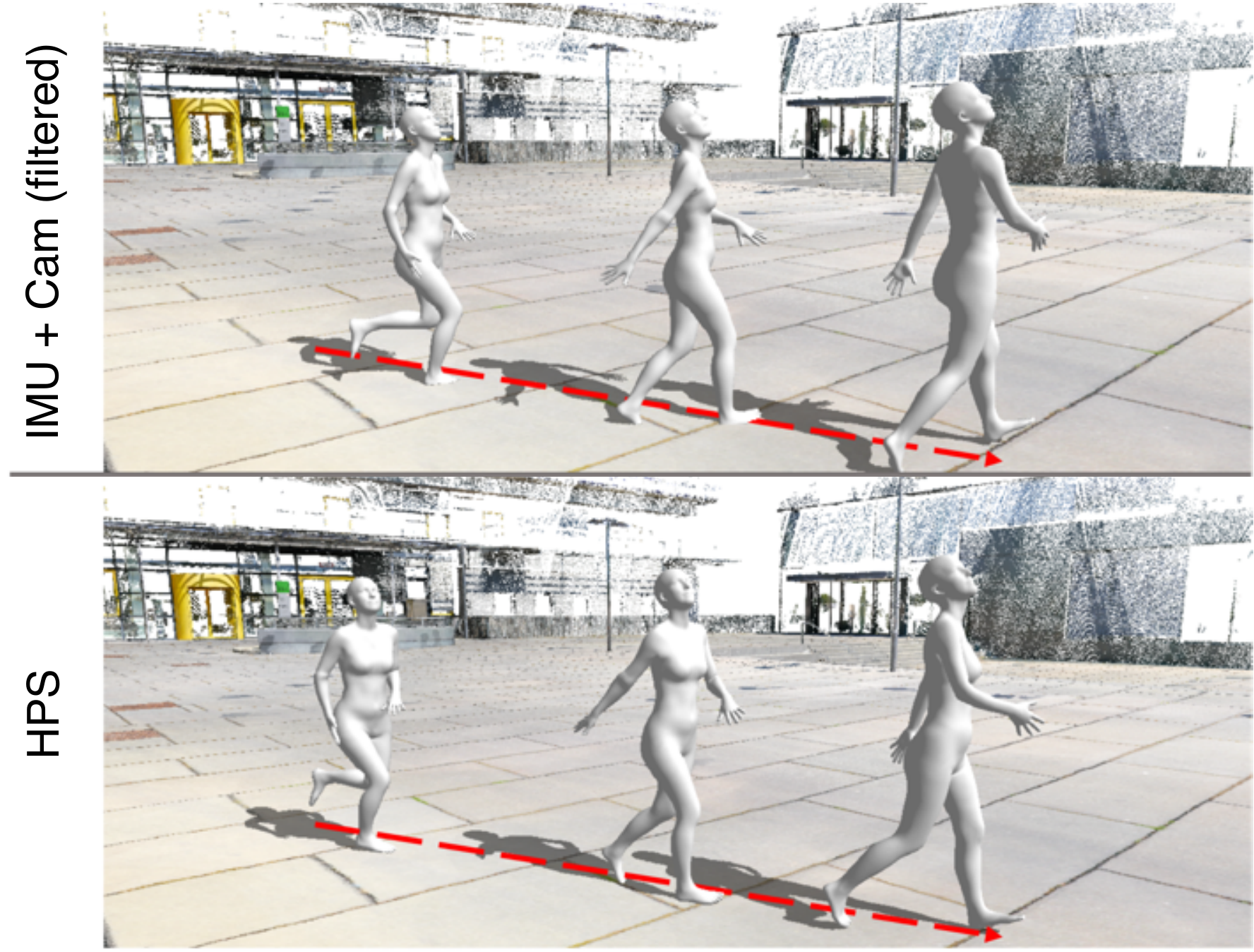}
    \caption{\textbf{Global body orientation improvement.} Combining the IMU pose with position from camera localization (IMU+Cam (filtered)) results in unnatural motion--the global body orientation does not face the direction of movement. By contrast, HPS correctly estimates the the global orientation. We refer to the video at project page~\cite{hps_project_page} for more visual examples.}
    \label{fig:optim_improvement_qualitativ}
    \vspace{-3mm}
\end{figure}

\textbf{Drift and Outliers:} 
In Tables \ref{tab:eval_table_apose} and \ref{tab:eval_table_variousact}, we compare HPS to the baselines. We observe that the IMU-only method drifts over time, particularly the global body translation and orientation. IMU+Cam corrects drift with camera localization, but produces translation noise and severe jitter. IMU+Cam (filtered) mitigates this, but lacks precision and suffers from global orientation errors (Fig.~\ref{fig:optim_improvement_qualitativ}). HPS w\textbackslash o scene further improves results, but without knowledge about foot-scene contacts, it is easily misled by incorrect camera localization, and the subject penetrates or flies over the ground. HPS results satisfy these scene constraints, and consistently achieve the best accuracy. HPS is inaccurate when filtered camera localization fails for a long period (see 2nd and 4th rows of Table~\ref{tab:eval_table_variousact}), but it can recover once the camera can be well localized in nearby frames (see 3rd row of Table~\ref{tab:eval_table_variousact}). Overall, the analysis reveals that HPS does not drift (error does not increase with distance traveled or time), and is robust to non-persistent camera localization outliers.

For scenes with with persistent camera localization failures (outdoor scenes, indoor scenes with repetitive patterns), we implemented a slightly modified version of HPS, described in the supplementary.

\textbf{Foot contacts:} We also report in Table~\ref{tab:eval_table_feet} the average foot-to-scene distance and foot-sliding-along-the-surface distance during contacts detected with the IMUs. HPS better preserves foot contact with the surface than the baselines, and has slightly lower foot-sliding compared to the raw IMU tracker, which also integrates constraints with a \emph{virtual} imaginary ground. 
Foot contacts in HPS result in stable and natural motion, see Fig.~\ref{fig:3D_contacts_qualitativ}, and the video~\cite{hps_project_page}.
\vspace{-8mm}

\begin{center}
\begin{figure*}[t!]
\setlength\tabcolsep{0.3pt}
\begin{minipage}{0.32\linewidth}
\centering
\includegraphics[width =\linewidth]{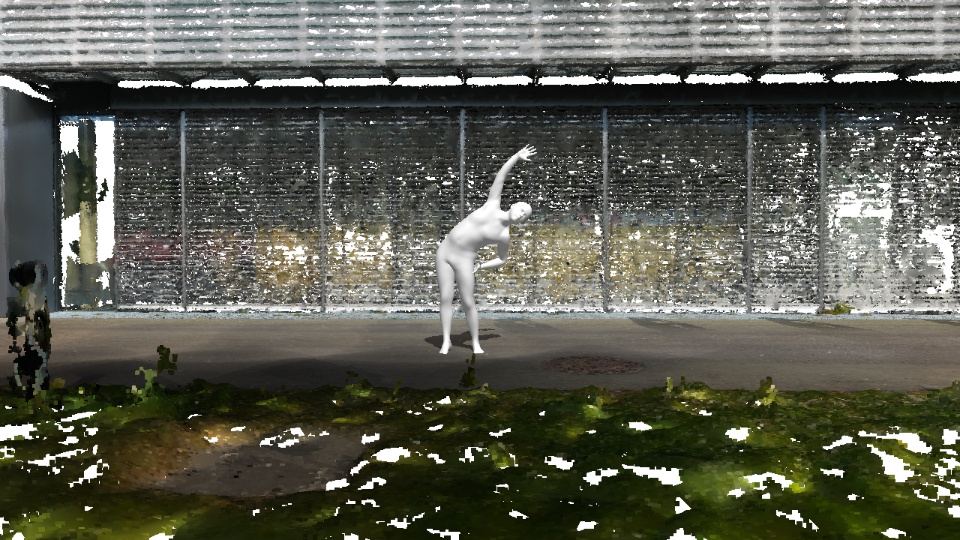}
\\
\includegraphics[width =\linewidth]{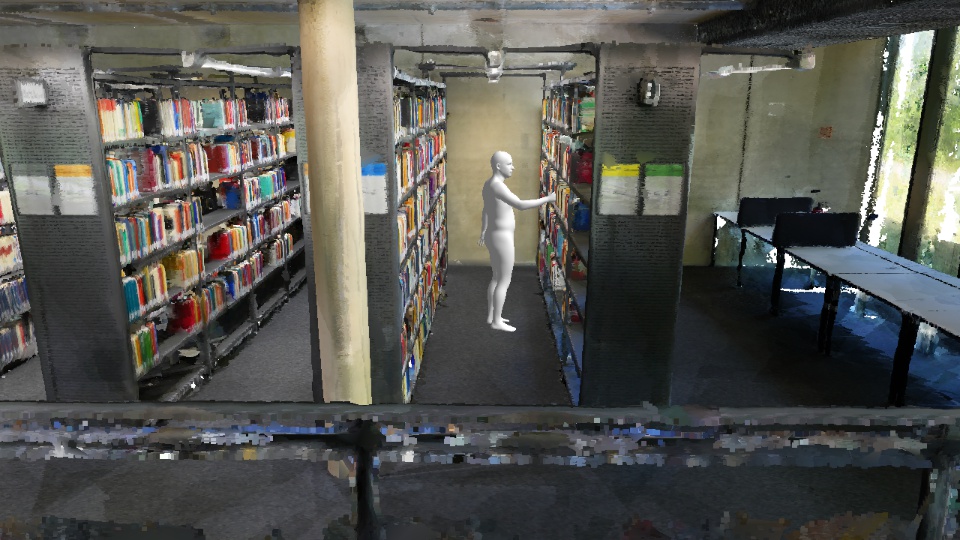}
\\ 
\includegraphics[width =\linewidth]{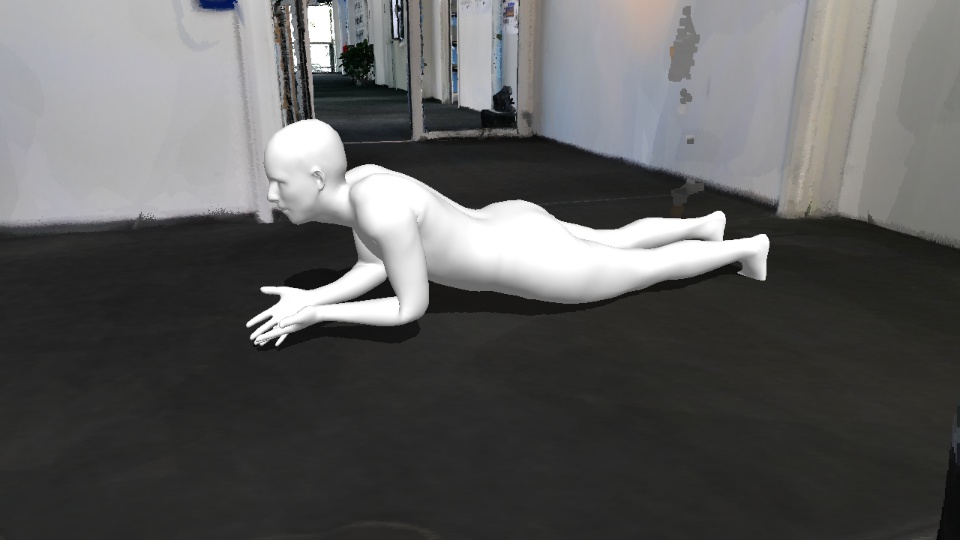}
\end{minipage}\hfill
\begin{minipage}{0.32\linewidth}
\includegraphics[width =\linewidth]{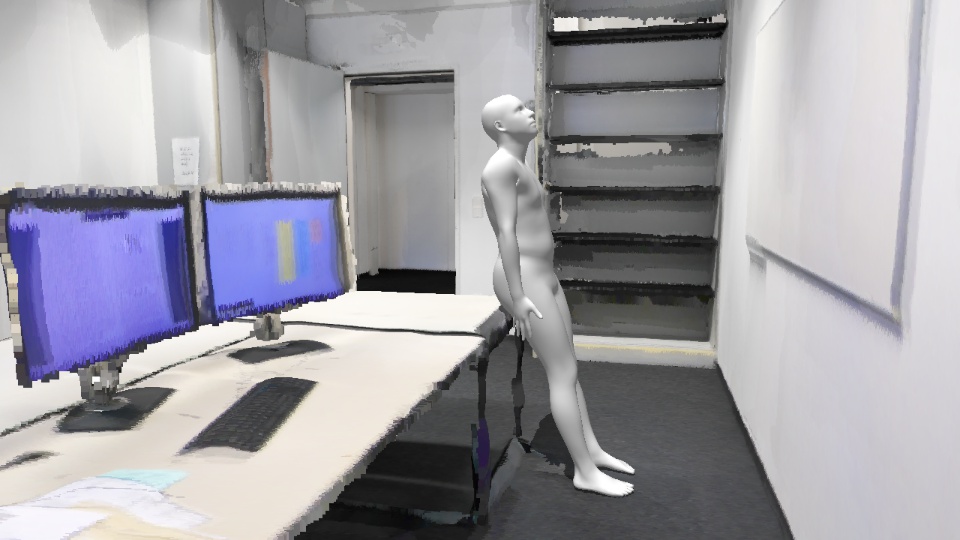}
\\ 
\includegraphics[width =\linewidth]{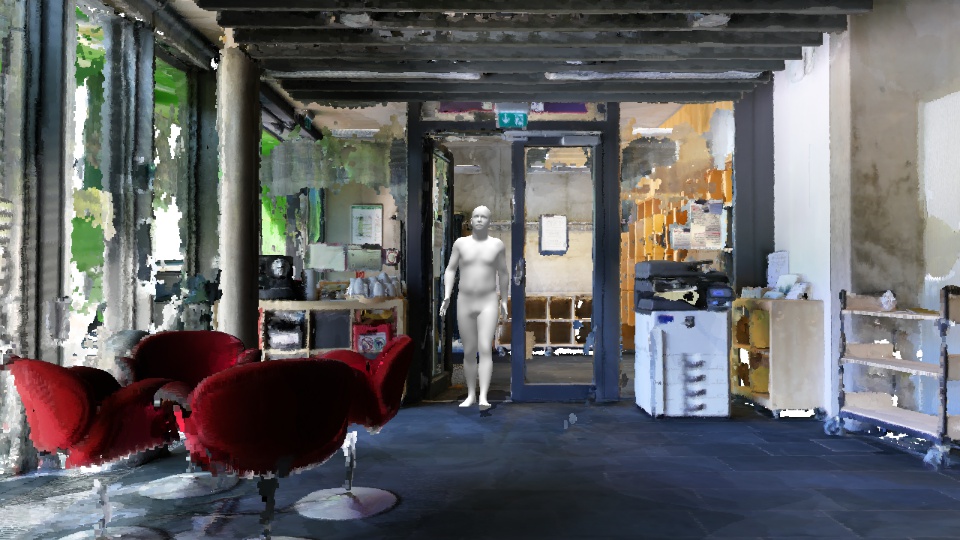}
\\ 
\includegraphics[width =\linewidth]{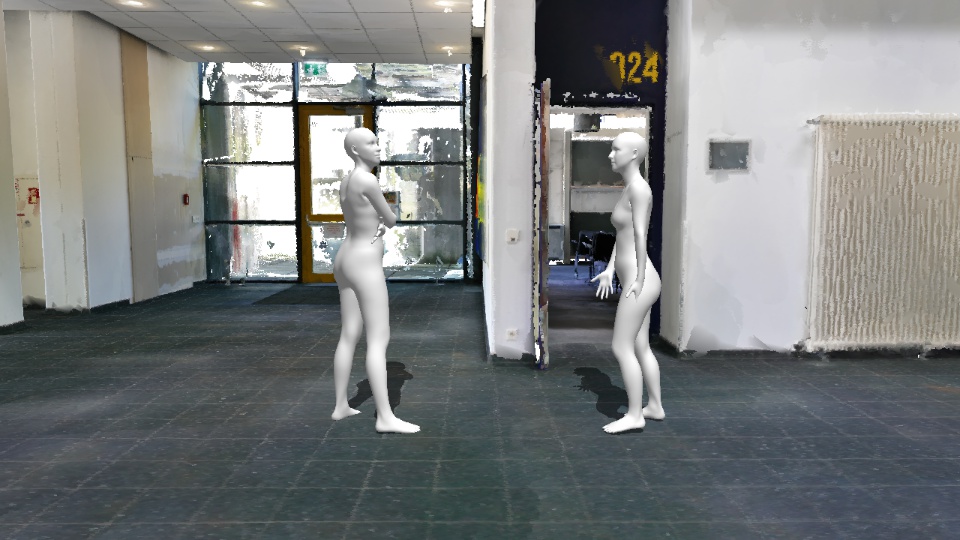}
\end{minipage}\hfill
\begin{minipage}{0.32\linewidth}
\includegraphics[width =\linewidth]{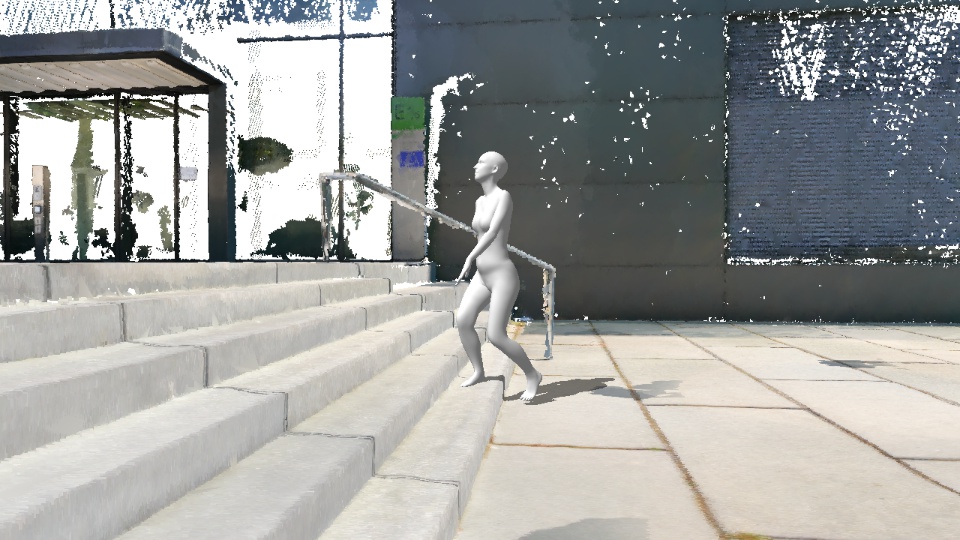}
\\ 
\includegraphics[width =\linewidth]{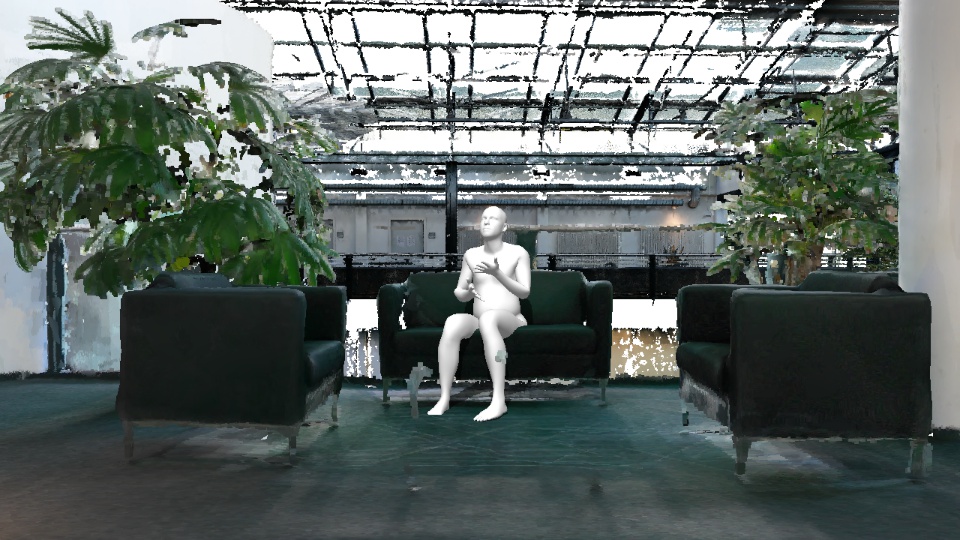}
\\ 
\includegraphics[width =\linewidth]{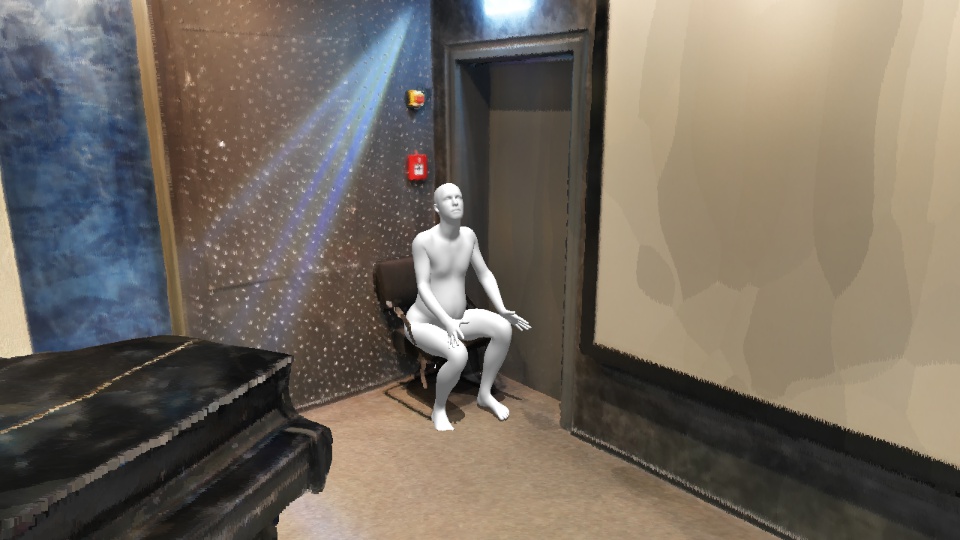}
\end{minipage}
\vspace{1mm}
\caption{\textbf{We show qualitative results of our method.} Our method can localize and estimate the 3D pose of people performing activities as diverse as exercising, dancing, reading, sitting, eating, talking in a range of indoor and outdoor scenes, all \emph{without} external cameras.}
\vspace{-3mm}

\label{fig:exp_qulitative_results}
\end{figure*}
\end{center}

\subsection{Qualitative evaluation}
\label{subsec:exp_qualit_eval}
In Fig.~\ref{fig:3D_contacts_qualitativ} we show the effect of foot contact constraints.
As we encourage contact with the scene surface each time a contact is detected, the human mesh does not fly in the air or penetrate the ground like the baseline. The motion is more stable and physically correct.
In Fig.~\ref{fig:exp_qulitative_results} we show examples of humans performing different actions including sitting, leaning on a table, dancing or performing push-ups. For more examples, please see the video at our project page~\cite{hps_project_page}.

%% file: sections/conclusions.tex
We introduced HPS, to the best of our knowledge, the first method to estimate full body pose registered with a pre-scanned 3D environment from \emph{only wearable sensors}.
We demonstrate that HPS produces natural human motion, removes the typical drift of pure IMU based systems, and is robust to non-persistent camera localization outliers. HPS is able to continuously track humans in large scenes ($300-1000m^2$) including multiple rooms and outdoors. 

The error of HPS does not accumulate with time or distance traveled. However, if camera localization is inaccurate for long periods of time, HPS performance deteriorates. This can be seen in the errors, which range from $4cm$ to $15cm$. Two factors influence localization accuracy: 1) Lack of features, 2) scene changes between the static 3D scan and the real images, captured from the head camera. 

While HPS achieves a remarkable accuracy and stability, many applications will require errors in localization and pose of less than $1cm$. We envision many exciting research directions to improve HPS. First, a local map could be built on the fly to update the large static scene with objects that move, and adding new objects. This would improve localization and allow interaction with dynamic objects.
It is not inconceivable that, in the future, a dynamic 3D reconstruction of the world will be stored on the cloud, and will be continuously updated from cameras worn by people~\cite{Aria}.
Second, camera localization could incorporate semantics~\cite{bloesch2018codeslam,zhi2019scenecode}, e.g. detecting static and reliable objects. 
Third, while HPS integrates foot contacts, scene constraints with other body parts can further improve results. More powerful would be to learn a model to \emph{anticipate human intent} to improve tracking. For example, we could detect when the person is about to sit on a chair, or about to grab an object. Conversely, HPS can be used to build models of environment interaction and navigation~\cite{habitat19iccv,xia2018gibson} from human captures consisting of several hours, as we believe natural behavior arises only during long recordings. Fourth, we want to combine HPS with virtual humans of appearance~\cite{patel20tailornet,bhatnagar2019mgn,mir20pix2surf,bhatnagar2020ipnet} to generate realistic data for training and evaluation of 3D human analysis methods. 

HPS is the first step in a new exciting research direction. 
We will \emph{release the HPS dataset and code} for research use~\cite{hps_project_page}, and hope it will foster new methods to perceive and model scenes and humans from an ego-centric perspective.  